\documentclass[runningheads]{llncs}
\usepackage{graphicx}
\usepackage{amsmath,amssymb} 
\usepackage{color}

\usepackage{booktabs}
\usepackage{multirow,bigdelim}

\begin{document}
\title{Question-Guided Hybrid Convolution for Visual Question Answering} 

\titlerunning{Question-Guided Hybrid Convolution for Visual Question Answering}
%
\author{Peng Gao\inst{1} \and
Pan Lu\inst{1} \and
Hongsheng Li \inst{1} \thanks{corresponding author} \and
Shuang Li\inst{1} \and
Yikang Li\inst{1} \and
Steven C.H. Hoi\inst{2} \and
Xiaogang Wang\inst{1}}
%
\authorrunning{Peng Gao, Pan Lu, Hongsheng Li, Shuang Li,  Yikang Li, Steven C.H.Hoi}
%

\institute{CUHK-SenseTime Joint Lab, The Chinese University of Hong Kong 
\email{\{penggao,hsli,sli,plu,ykli,xgwang\}@ee.cuhk.edu.hk} \and
School of Information Systems, Singapore Management Univeristy\\
\email{chhoi@smu.edu.sg}}
\maketitle              
\begin{abstract}
In this paper, we propose a novel Question-Guided Hybrid Convolution (QGHC) network for Visual Question Answering (VQA). Most state-of-the-art VQA methods fuse the high-level textual and visual features from the neural network and abandon the visual spatial information when learning multi-modal features.
To address these problems, question-guided kernels generated from the input question are designed to convolute with visual features for capturing the textual and visual relationship in the early stage. The question-guided convolution can tightly couple the textual and visual information but also introduce more parameters when learning kernels. We apply the group convolution, which consists of question-independent kernels and question-dependent kernels, to reduce the parameter size and alleviate over-fitting.
The hybrid convolution can generate discriminative multi-modal features with fewer parameters.
The proposed approach is also complementary to existing bilinear pooling fusion and attention based VQA methods. By integrating with them, our method could further boost the performance. Experiments on VQA datasets validate the effectiveness of QGHC. 

\end{abstract}
\section{Introduction}

\begin{figure}[t]
        \begin{center}
                \includegraphics[width=0.8\linewidth]{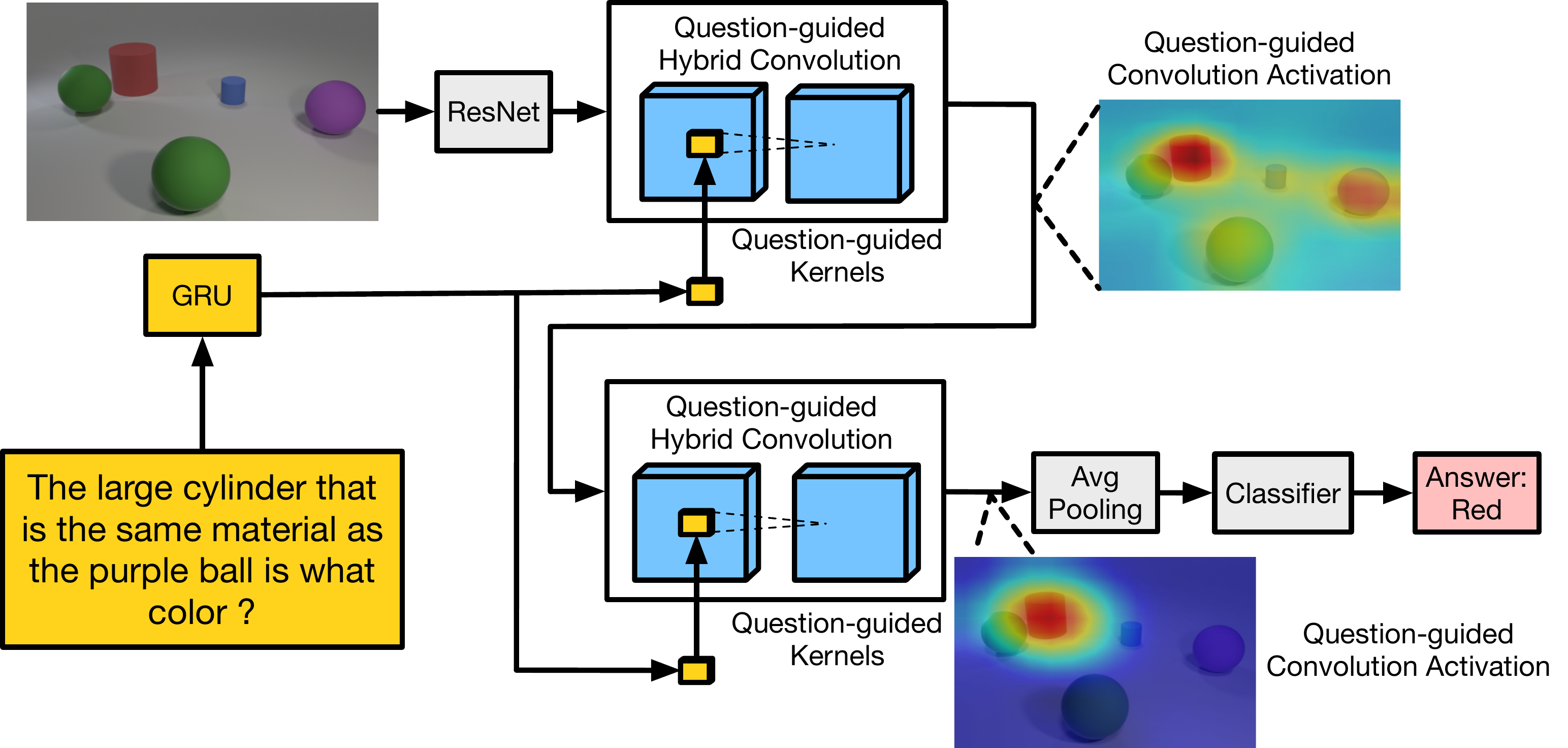}
        \end{center}
        \caption{Illustration of using multiple Question-guided Hybrid Convolution modules for VQA. Question-guided kernels are predicted by the input question and convoluted with visual features. Visualization of the question-guided convolution activations show they gradually focus on the regions corresponding to the correct answer.}
        \label{fig:QGC}
\end{figure}

Convolution Neural Networks (CNN) \cite{krizhevsky2012imagenet} and Recurrent Neural Networks (RNN) \cite{sutskever2014sequence} have shown great success in vision and language tasks. 
Recently, CNN and RNN are jointly trained for learning feature representations for multi-modal tasks, including image captioning \cite{li2017scene,xu2015show}, text-to-image retrieval \cite{hu2016natural,li2017person}, 
and Visual Question Answering (VQA) \cite{antol2015vqa,kim2016hadamard,ben2017mutan,lu2018rvqa}. 
Among the vision-language tasks, VQA is one of the most challenging problems. Instead of embedding images and their textual descriptions into the same feature subspace as in the text-image matching problem \cite{frome2013devise,reed2016learning,li2017identity}, VQA requires algorithms to answer natural language questions about the visual contents. The methods are thus designed to understand both the questions and the image contents to reason the underlying truth.

To infer the answer based on the input image and question, it is important to fuse the information from both modalities to create joint representations. Answers could be predicted by learning classifiers on the joint features. Early VQA methods \cite{zhou2015simple} fuse textual and visual information by feature concatenation. State-of-the-art feature fusion methods, such as Multimodal Compact Bilinear pooling (MCB) \cite{fukui2016multimodal}, utilize bilinear pooling to learn multi-model features.

However, the above type of methods have main limitations. The 
multi-modal features are fused in the latter model stage and the spatial information from visual features gets lost before feature fusion. 
The visual features are usually obtained by averaging the output of the last pooling layer and represented as 1-d vectors. But such operation abandons the spatial information of input images. 
In addition, the textual and visual relationship is modeled only on the topmost layers and misses details from the low-level and mid-level layers. 

To solve these problems, we propose a feature fusion scheme that generates multi-modal features by applying question-guided convolutions on the visual features (see Figure \ref{fig:QGC}). The mid-level visual features and language features are first learned independently using CNN and RNN. The visual features are designed to keep the spatial information. And then a series of kernels are generated based on the language features to convolve with the visual features.
Our model tightly couples the multi-modal features in an early stage to better capture the spatial information before feature fusion. 
One problem induced by the question-guided kernels is that the large number of parameters make it hard to train the model. Directly predicting ``full'' convolutional filters requires estimating thousands of parameters (\textit{e.g.} $256$ number of $3\times3$ filters convolve with the 256-channel input feature map). This is memory-inefficient and time-consuming, and does not result in satisfactory performances (as shown in our experiments).

Motivated by the group convolution \cite{chollet2016xception,krizhevsky2012imagenet,xie2016aggregated}, we decompose large convolution kernels into group kernels, each of which works on a small number of input feature maps. In addition, only a portion of such group convolution kernels (\emph{question-dependent kernels}) are predicted by RNN and the remaining kernels (\emph{question-independent kernels}) are freely learned via back-propagation. Both question-dependent and question-independent 
kernels are shown to be important, and we name the proposed operation as \emph{Question-guided Hybrid Convolution (QGHC)}. The visual and language features are deeply fused to generate discriminative multi-modal features. The spatial relations between the input image and question could be well captured by the question-guided convolution. Our experiments on VQA datasets validate the effectiveness of our approach and show advantages of the proposed feature fusion over the state-of-the-arts.

Our contributions can be summarized in threefold. 1) We propose a novel multi-modal feature fusion method based on question-guided convolution kernels. The relative visual regions have high response to the input question and spatial information could be well captured by encoding such connection in the QGHC model. The QGHC explores deep multi-modal relationships which benefits the visual question reasoning.
2) To achieve memory efficiency and robust performance in the question-guided convolution, we propose the group convolution to learn kernel parameters. The question-dependent kernels model the relationship of visual and textual information while the question-independent kernels reduce parameter size and alleviate over-fitting.
3) Extensive experiments and ablation studies on the public datasets show the effectiveness of the proposed QGHC and each individual component. Our approach outperforms the state-of-the-art methods using much fewer parameters.

\section{Related work}
\textbf{Bilinear pooling for VQA.} Solving the VQA problem requires the algorithms to understand the relation between images and questions. It is important to obtain discriminative multi-modal features for accurate answer prediction. Early methods utilize feature concatenation \cite{zhou2015simple} for multi-modal feature fusion \cite{lin2015bilinear,li2017identity,li2017person}. 
Recently, bilinear pooling methods are introduced for VQA to capture high-level interactions between visual and textual features. Multimodal Compact Bilinear Pooling (MCB) \cite{fukui2016multimodal} projects the language and visual features into a higher dimensional space and convolves them in the Fast Fourier Transform space. 
In Multimodal Low-rank Bilinear (MLB) \cite{kim2016hadamard}, the weighting tensor for bilinear pooling is approximated by three weight matrices, which enforces the rank of the weighting tensor to be low-rank. The multi-modal features are obtained as the Hadamard product of the linear-projected visual and language features. Ben-younes \textit{et al} \cite{ben2017mutan} propose the Multimodal Tucker Fusion (MUTAN), which unifies MCB and MLB into the same framework . The weights are decomposed according to the Tucker decomposition. MUTAN achieves better performance than MLB and MCB with fewer parameters. 

\textbf{Attention mechanisms in language and VQA tasks.} The attention mechanisms \cite{xu2016ask,li2018diversity} are originally proposed for solving language-related tasks \cite{bahdanau2014neural}. Xu \textit{et al} \cite{xu2016ask} introduce an attention mechanism for image captioning, which shows that the attention maps could be adaptively generated for predicting captioning words. Based on \cite{xu2016ask}, Yang \textit{et al} \cite{yang2016stacked} propose to stack multiple attention layers so that each layer can focus on different regions adaptively. In \cite{lu2016hierarchical}, a co-attention mechanism is proposed. The model generates question attention and spatial attention masks so that salient words and regions could be jointly selected for more effective feature fusion. 
Similarly, Lu \textit{et al} \cite{lu2018coatt} employ a co-attention mechanism to simultaneously learn free-form and detection-based image regions related to the input question.
In MCB \cite{fukui2016multimodal}, MLB \cite{kim2016hadamard}, and MUTAN \cite{ben2017mutan}, attention mechanisms are adopted to partially recover the spatial information from the input image. Question-guided attention methods \cite{chen2015abc,xu2016ask} are proposed to generate attention maps from the question. 

\textbf{Dynamic Network.} Network parameters could be dynamically predicted across different modalities.
Our approach is mostly related to methods in this direction. In \cite{noh2016image}, language are used to predict parameters of a fully-connected (FC) layer for learning visual features. However, the predicted fully-connected layer cannot capture spatial information of the image. To avoid introducing too many parameters, they predict only a small portion of parameters using a hashing function. However, this strategy introduces redundancy because the FC parameters only contain a small amount of training parameters. In \cite{de2017modulating}, language is used to modulate the mean and variance parameters of the Batch Normalization layers in the visual CNN. However, learning the interactions between two modalities by predicting the BN parameters has limited learning capacity. 
We conduct comparisons with \cite{noh2016image} and \cite{de2017modulating}. Our proposed method shows favorable performance. We notice that \cite{li2017tracking} use language-guided convolution for object tracking. However, they predict all the parameters which is difficult to train.

\textbf{Group convolution in deep neural networks.} 
Recent research found that the combination of depth-wise convolution and channel shuffle with group convolution could reduce the number of parameters in CNN without hindering the final performance. Motivated by Xception \cite{chollet2016xception}, ResNeXt \cite{xie2016aggregated}, and ShuffleNet \cite{zhang2017shufflenet}, we decompose the visual CNN kernels into several groups. By shuffling parameters among different groups, our model can reduce the number of predicted parameters and improve the answering accuracy simultaneously. Note that for existing CNN methods with group convolution, the convolutional parameters are solely learned via back-propagation. In contrast, our QGHC consists of question-dependent kernels that are predicted based on language features and question-independent kernels that are freely updated.

\section{Visual Question Answering with Question-guided Hybrid Convolution}
\label{sec:approach}

ImageQA systems take an image and a question as inputs and output the predicted answer for the question. ImageQA algorithms mostly rely on deep learning models and design effective approaches to fuse the multi-modal features for answering questions. Instead of fusing the textual and visual information in high level layers, such as feature concatenation in the last layer, we propose a novel multi-modal feature fusion method, named Question-guided Hybrid Convolution (QGHC). Our approach couples the textual-visual features in early layers for better capturing textual-visual relationships. It learns question-guided convolution kernels and reserves the visual spatial information before feature fusion, and thus achieves accurate results.
The overview of our method is illustrated in Figure \ref{fig:QGC}. The network predicts convolution kernels based on the question features, and then convolve them with visual feature maps. We stack multiple question-guided hybrid convolution modules, an average pooling layer, and a classifier layer together. The output of the language-guided convolution is the fused textual-visual features maps which used for answering questions. To improve the memory efficiency and experimental accuracy, we utilize the group convolution to predict a portion of convolution kernels based on the question features.

\subsection{Problem formulation}
\label{sec:problem}

Most state-of-the-art VQA methods rely on deep neural networks for learning discriminative features of the input image $I$ and question $q$. Usually, Convolutional Neural Networks (CNN) are adopted for learning visual features, while Recurrent Neural Networks (RNN) (\textit{e.g.}, Long Short-Term Memory (LSTM) or Gated Recurrent Unit (GRU)) encode the input question, \textit{i.e.},
\begin{align}
  f_v &= \textnormal{CNN} (I; \theta_v), \label{eq:visual} \\
  f_q &= \textnormal{RNN} (q; \theta_q), \label{eq:language}
\end{align}
where $f_v$ and $f_q$ represent visual features and question features respectively.

Conventional ImageQA systems focus on designing robust feature fusion functions to generate multi-modal image-question features for answer prediction.
Most state-of-the-art feature fusion methods fuse 1-d visual and language feature vectors in a symmetric way to generate the multi-modal representations. The 1-d visual features are usually generated by the deep neural networks (\textit{e.g.}, GoogleNet and ResNet) with a global average pooling layer. Such visual features $f_v$ and the later fused textual-visual features abandon spatial information of the input image and thus less robust to spatial variations. 

\subsection{Question-guided Hybrid Convolution (QGHC) for multi-modal feature fusion}
\label{sec:problem1}

To fully utilize the spatial information of the input image, we propose Language-guided Hybrid Convolution for feature fusion. Unlike bilinear pooling methods that treat visual and textual features in a symmetric way, our approach performs the convolution on visual feature maps and the convolution kernels are predicted based on the question features which can be formulated as:
\begin{align}
  f_{v+q} &= \textnormal{CNN}_p ( I; \tilde{\theta}_v(f_q)) \label{eq:naive},
\end{align}
where $\textnormal{CNN}_p$ is the output before the last pooling layer, $\tilde{\theta}_v(f_q)$ denotes the convolutional kernels predicted based on the question feature $f_q \in \mathbb{R}^d$, and the convolution on visual feature maps with the predicted kernels $\tilde{\theta}_v(q)$ results in the multi-modal feature maps $f_{v+q}$.

However, the naive solution of directly predicting ``full'' convolutional kernels is memory-inefficient and time-consuming. Mapping the question features to generate full CNN kernels contains a huge number of learnable parameters. In our model, we use the fully-connected layer to learn the question-guided convolutional kernels. To predict a commonly used $3\times 3 \times 256 \times 256$ kernel from a 2000-d question feature vector, the FC layer for learning the mapping generates 117 million parameters, which is hard to learn and causes over-fitting on existing VQA datasets. In our experiments, we validate that the performance of the naive solution is even worse than the simple feature concatenation.

To mitigate the problem, we propose to predict parameters of group convolution kernels. The group convolution divides the input feature maps into several groups along the channel dimension, and thus each group has a reduced number of channels for convolution. Outputs of convolution with each group are then concatenated in the channel dimension to produce the output feature maps. In addition, we classify the convolution kernels into dynamically-predicted kernels and freely-updated kernels. The dynamic kernels are question-dependent, which are predicted based on the question feature vector $f_q$. The freely-updated kernels are question-independent. They are trained as conventional convolution kernels via back-propagation. The dynamically-predicted kernels fuse the textual and visual information in early model stage which better capture the multi-model relationships. The freely-updated kernels reduce the parameter size and ensure the model can be trained efficiently. By shuffling parameters among these two kinds of kernels, our model can achieve both the accuracy and efficiency.
During the testing phase, the dynamic kernels are decided by the questions while the freely updated kernels are fixed for all input image-question pairs.

Formally, we substitute Eqn. (\ref{eq:naive}) with the proposed QGHC for VQA,
\begin{align}
  f_{v+q} &= \textnormal{CNN}_g \left( I; \tilde{\theta}_v(f_q), \theta_v \right),\\
  \widehat{a} &= \textnormal{MLP}(f_{v+q}),
\end{align}
where CNN$_g$ denotes a group convolution network with dynamically-predicted kernels $\tilde{\theta}_v(f_q)$ and freely-updated kernels $\theta_v$. The output of the CNN $f_{v+q}$ fuses the textual and visual information and infers the final answers. MLP is a multi-layer perception module and $\widehat{a}$ is the predicted answers.

The freely-updated kernels can capture pre-trained image patterns and we fix them during the testing stage. The dynamically-predicted kernels are dependent on the input questions and capture the question-image relationships. Our model fuses the textual and visual information in early model stage by the convolution operation. The spatial information between two modalities is well preserved which leads to more accurate results than previous feature concatenation strategies. The combination of the dynamic and freely-updated kernels is crucial important in keeping both the accuracy and efficiency and shows promising results in our experiments.

\subsection{QGHC module}
\label{sec:lghc_module}

\begin{figure}[t]
  \begin{center}
    \includegraphics[width=0.6\linewidth]{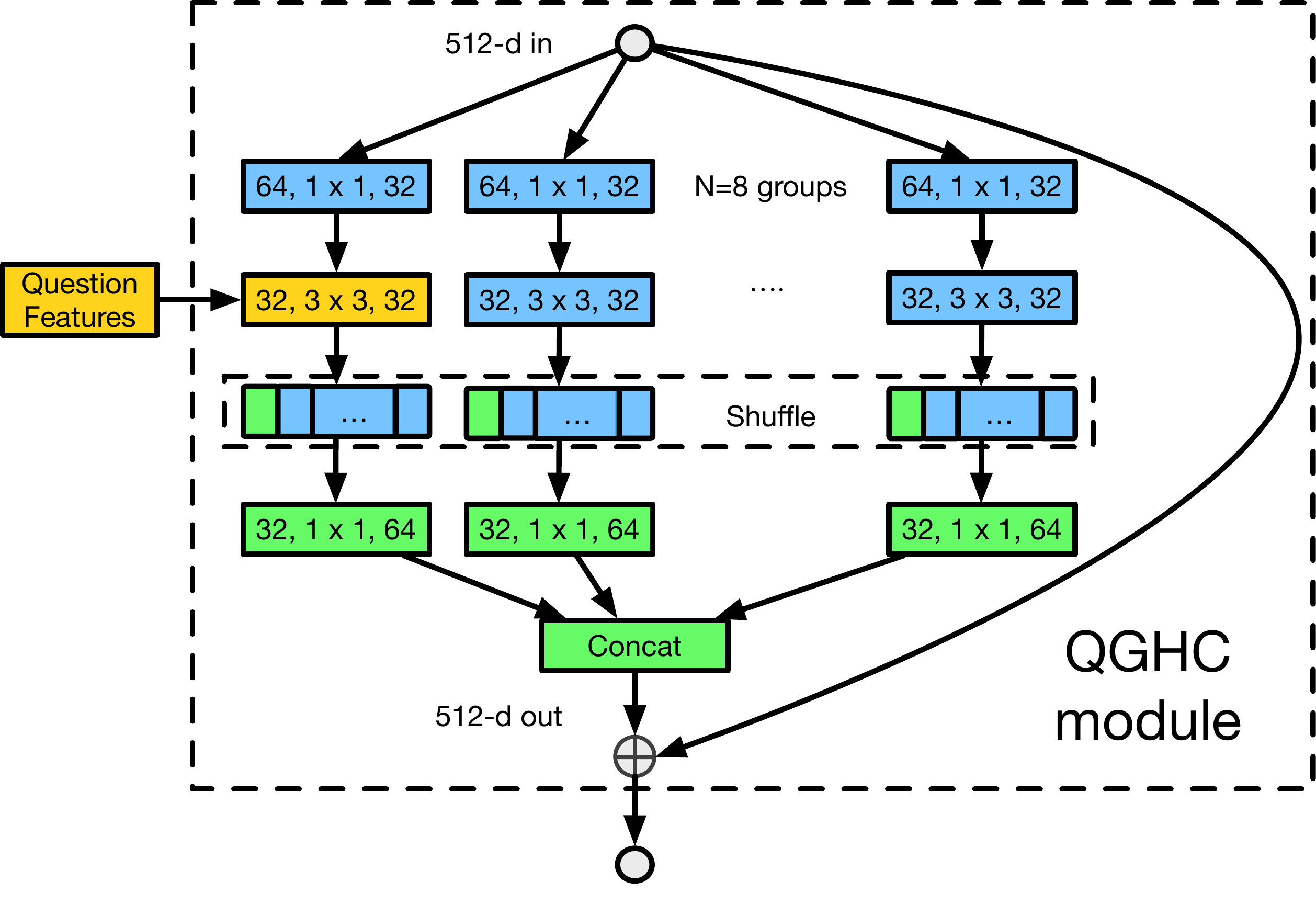}
  \end{center}
  \caption{Network structure of our QGHC module with $N=8$ and $C_i=C_o=512$. The question features are used to learn $n$ convolution groups in the $3\times 3$ convolution layer (the yellow block). A group shuffling layer is utilized to share the textual information from question-guided kernels to the whole network.}
  \label{fig:QGHC}
\end{figure}

We stack multiple QGHC modules to better capture the interactions between the input image and question. Inspired by ResNet \cite{he2016deep} and ResNeXt \cite{xie2016aggregated}, our QGHC module consists of a series of $1\times 1$, $3\times 3$, and $1\times 1$ convolutions. 

As shown in Figure \ref{fig:QGHC}, the module is designed similarly to the ShffuleNet \cite{zhang2017shufflenet} module with group convolution and identity shortcuts. The $C_i$-channel input feature maps are first equally divided into $N$ groups (paths). Each of the $N$ groups then goes through 3 stages of convolutions and outputs $C_o/N$-d feature maps. For each group, the first convolution is a $1\times 1$ convolution that outputs $C_i /2N$-channel feature maps. The second $3\times 3$ convolution outputs $C_i /2N$-channel feature maps, and the final $1\times 1$ convolution outputs $C_o/N$-channel feature maps. We add a group shuffling layer after the $3\times 3$ convolution layer to make features between different groups interact with each other and keep the advantages of both the dynamically-predicted kernels and freely-updated kernels. The output of $C_o/N$-channel feature maps for the $N$ groups are then concatenated together along the channel dimension. For the shortcut connection, a $1\times 1$ convolution transforms the input feature maps to $C_o$-d features, which are added with the output feature maps. Batch Normalization and ReLU are performed after each convolutional operation except for the last one, where ReLU is performed after the addition with the shortcut.

The $3 \times 3$ group convolution is guided by the input questions. We randomly select $n$ group kernels. Their parameters are predicted based on the question features.
Those kernel weights are question-dependent and are used to capture location-sensitive question-image interactions. 
The remaining $N-n$ group kernels have freely-updated kernels. They are updated via back-propagation in the training stage and are fixed for all images during testing. These kernels capture the pre-trained image patterns or image-question patterns. They are constant to the input questions and images.

\begin{figure*}[t!]
        \begin{center}
                \includegraphics[width=0.8\linewidth]{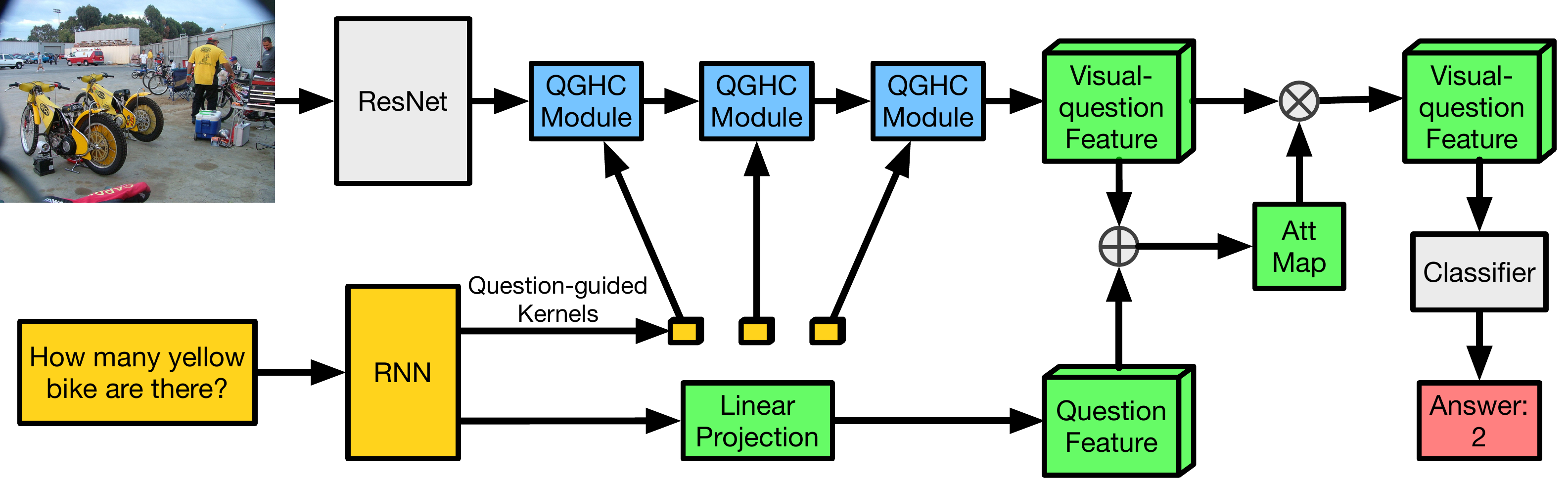}
        \end{center}
        \caption{The proposed QGHC network with three stacked QGHC modules for VQA. Question-guided kernels are learned based on the input question and convoluted with visual feature maps to generate multi-modal features for the answer prediction.}
        \label{fig:overall}
\end{figure*}

\subsection{QGHC network for visual question answering}
\label{sec:lghc_network}

The network structure for our QGHC network is illustrated in Figure \ref{fig:overall}. The ResNet \cite{he2016deep} is first pre-trained on the ImageNet to extract mid-level visual features. The question features are generated by a language RNN model.

The visual feature maps are then send to three QGHC modules with $N=8$ groups and $C_o=512$. The output of the QGHC modules $f_{v+q}$ has the same spatial sizes with the input feature maps. A global average pooling is applied to the final feature maps to generate the final multi-modal feature representation for predicting the most likely answer $\widehat{a}$.

To learn the dynamic convolution kernels in the QGHC modules, the question feature $f_q$ is transformed by two FC layers with a ReLU activation in between. The two FC layers first project the question to a $9216$-d vector. The $3\times 3$ question-dependent kernel weights of the three QGHC modules are obtained by reshaping the learned parameters into $3\times 3 \times 32 \times 32$. However, directly training the proposed network with both dynamically-predicted kernels and freely-updated kernels is non-trivial. The dynamic kernel parameters are the output of the ReLU non-linear function with different magnitudes compared with the freely-updated kernel parameters. We adopt the Weight Normalization \cite{salimans2016weight} to balance the weights between the two types of $3 \times 3$ kernels, which stabilizes the training of the network.

\subsection{QGHC network with bilinear pooling and attention}
\label{ssec:attention}

Our proposed QGHC network is also complementary with the existing bilinear pooling fusion methods and the attention mechanism.

To combine with the MLB fusion scheme \cite{kim2016hadamard}, the multi-modal features extracted from the global average pooling layer could be fused with the RNN question features again using a MLB. The fused features could be used to predict the final answers. The second stage fusion of textual and visual features brings a further improvement on the answering accuracy in our experiments.

We also apply an attention model to better capture the spatial information. The original global average pooling layer is thus replaced by the the attention map. 
To weight more on locations of interest, a weighting map is learned by attention mechanism. A $1\times 1$ convolution following a spatial Softmax function generates the attention weighting map. The final multi-modal features is the weighted summation of features at all the locations. The output feature maps from the last QGHC module are added with the linearly transformed question features. The attention mechanism is shown as the green rectangles in Figure \ref{fig:overall}. 

\section{Experiments}

We test our proposed approach and compare it with the state-of-the-arts on two public datasets, the CLEVR dataset \cite{johnson2016clevr} and VQA dataset \cite{antol2015vqa}.

\begin{table}[tb]
\scriptsize
\centering
\begin{tabular}{l c c c c c}
\multirow{2}{*}{Model} &      \multicolumn{2}{c}{Parameter size} & \multicolumn{1}{c}{val} \\
\cmidrule(lr){2-3} \cmidrule(lr){4-4} 
 &  QD Weights & QI Weights  & All\\
\hline
QGHC &  5.4M  & 0.9M  & 59.24\\
\hline
QGHC-1 & 1.8M & 0.3M   & 58.88\\
QGHC-2 & 3.6M   & 0.6M   & 59.04\\
QGHC-4 & 7.2M  & 1.2M   & 59.13\\
QGHC-1/2 &  1.3M   & 0.7M    & 58.78\\
\hline
QGHC-group 4 & 8.7M  & 2.1M    & 59.01  \\
QGHC-group 16 & 1.3 M   & 0.15M    & 58.22  \\
QGHC-w/o shuffle & 5.4M & 0.9M  & 58.92  \\
\hline
QGHC-1-naive &  471M   &  0M    & 55.32 \\
QGHC-1-full &  117M  & 0.2M   & 57.01 \\
QGHC-1-group &  14M  & 0.03M     & 58.41 \\
\hline
QGHC+concat  &  - & -   & 59.80\\
QGHC+MUTAN  & - & -   & 60.13\\
QGHC+att. &  -  &  -   & 60.64 \\
\bottomrule
\end{tabular}
\caption{Ablation studies of our proposed QGHC network on the VQA dataset. QD and QI stands for question-dependent and -independent kernels.}
\label{tab:ablation}
\end{table}

\subsection{VQA Dataset}
\label{ssec:vqa}

\subsubsection{Data and experimental setup.}
\label{sssec:vqa_data}

The VQA dataset is built from 204,721 MS-COCO images with human annotated questions and answers. On average, each image has 3 questions and 10 answers for each question. The dataset is divided into three splits: training (82,783 images), validation (40,504 images) and testing (81,434 images).
A testing subset named \emph{test-dev} with 25\% samples can be evaluated multiple times a day. We follow the setup of previous methods and perform ablation studies on the testing subset.
Our experiments focus on the open-ended task, which predict the correct answer in the free-form language expressions.
If the predicted answer appears more than 3 times in the ground truth answers, the predicted answer would be considered as correct. 

Our models have the same setting when comparing with the state-of-the-art methods. The compared methods follow their original setup. For the proposed approach, images are resized to $448 \times 448$. The $14 \times 14 \times 2048$ visual features are learned by an ImageNet pre-trained ResNet-152, and the question is encoded to a 2400-d feature vector by the skip-thought \cite{kiros2015skip} using GRU. The candidate questions are selected as the most frequent 2,000 answers in the training and validation sets. The model is trained using the ADAM optimizer with an initial learning rate of $10^{-4}$. For results on the validation set, only the training set is used for training. For results on test-dev, we follow the setup of previous methods, both the training and validation data are used for training.

\subsubsection{Ablation studies on the VQA dataset.}
\label{sssec:ablation_vqa}
We conduct ablation studies to investigate factors that influence the final performance of our proposed QGHC network. The results are shown in Table \ref{tab:ablation}.
Our default QGHC network (denoted as \emph{QGHC}) has a visual ResNet-152 followed by three consecutive QGHC modules. Each QGHC module has a $1\times 1$ stage-1 convolution with freely-updated kernels, a $3\times 3$ stage-2 convolution with both dynamically-predicted kernels and freely-updated kernels, and another $1\times 1$ convolution stage with freely-updated kernels (see Figure \ref{fig:QGHC}). Each of these three stage convolutions has 8 groups. They have 32, 32, and 64 output channels respectively.

We first investigate the influence of the number of QGHC modules and the number of convolution channels. We list the results of different number of QGHC modules in Table \ref{tab:ablation}. \textit{QGHC-1}, \textit{QGHC-2}, \textit{QGHC-4} represent 1, 2, and 4 QGHC modules respectively. As shown in Table \ref{tab:ablation}, the parameter size improves as the number of QGHC increases but there is no further improvement when stacking more than 3 QGHC modules. We therefore keep 3 QGHC modules in our model.
We also test halving the numbers of output channels of the three group convolutions to 16, 16, and 32 (denoted as \emph{QGHC-1/2}). The results show that halving the number of channels only slightly decreases the final accuracy.

We then test different group numbers. We change the group number from 8 to 4 (\emph{QGHC-group 4}) and 16 (\emph{QGHC-group 16}). Our proposed method is not sensitive to the group number of the convolutions and the model with 8 groups achieves the best performance. 
We also investigate the influence of the group shuffling layer.
Removing the group shuffling layer (denoted as \emph{QGHC-w/o shuffle}) decreases the accuracy by 0.32\% compared with our model. The shuffling layer makes features between different groups interact with each other and is helpful to the final results.

For different QGHC module structures, we first test a naive solution. The QGHC module is implemented as a single $3 \times 3$ ``full'' convolution without groups. Its parameters are all dynamically predicted by question features (denoted as \emph{QGHC-1-naive}). We then convert the single $3 \times 3$ full convolution to a series of $1 \times 1$, $3 \times 3$, $1 \times 1$ full convolutions with residual connection between the input and output feature maps (denoted as \emph{QGHC-1-full}), where the $3\times 3$ convolution kernels are all dynamically predicted by the question features. The improvement of QGHC-1-full over QGHC-1-naive demonstrates the advantages of the residual structure. Based on QGHC-1-full, we convert all the full convolutions to group convolutions with 8 groups (denoted as \emph{QGHC-1-group}). The results outperforms QGHC-1-full, which show the effectiveness of the group convolution. 
However, the accuracy is still inferior to our proposed QGHC-1 with hybrid convolution. The results demonstrate that the question-guided kernels can help better fuse the textual and visual features and achieve robust answering performance. 

Finally, we test the combination of our method with different additional components. 1) The multi-modal features are concatenated with the question features, and then fed into the FC layer for answer prediction. (denoted as \emph{QGHC+concat}). It results in a marginal improvement in the final accuracy. 2) We use MUTAN \cite{ben2017mutan} to fuse our QGHC-generated multi-modal features with question features again for answer prediction (denoted as \emph{QGHC+MUTAN}). It has better results than QGHC+concat. 3) The attention is also added to QGHC following the descriptions in Section \ref{ssec:attention} (denoted as \emph{QGHC+att.}).

\begin{table}[tb]
\scriptsize
\centering
\begin{tabular}{l c c c c c c}
\toprule
\multirow{2}{*}{Model} & \multirow{2}{*}{\#parameters} &  \multicolumn{4}{c}{test-dev} & \multicolumn{1}{c}{val} \\
\cmidrule(lr){3-6} \cmidrule(lr){7-7} 
 &  & Y/N  & Number & Other & All & All\\
\hline
Concat \cite{zhou2016learning}& - &79.25 & 36.18 & 46.69 & 58.91 & 56.92 \\
MCB \cite{fukui2016multimodal}&32M &80.81 & 35.91 & 46.43 & 59.40 & 57.39 \\
MLB \cite{kim2016hadamard}&7.7M &82.02 & 36.61 & 46.65 & 60.08 & 57.91 \\
MUTAN \cite{ben2017mutan}&4.9M &81.45 & 37.32 & 47.17 & 60.17 & 58.16\\
MUTAN+MLB \cite{ben2017mutan}&17.5M &82.29 & 37.27 & 48.23 & 61.02 & 58.76\\
MFB \cite{yu2017multi}&-&81.80  & 36.70 &  51.20  & 62.20  &  -   \\
DPPNet \cite{noh2016image}&-&80.71 & 37.23 & 41.69 & 57.22 &   -    \\  
\hline
QGHC-1    &2.1M & -    &   -   &    -  &   -   &   58.88 \\
QGHC      &5.4M &82.39 & \textbf{37.36} & 53.24 & 63.48 & 59.24 \\
QGHC+concat &-&82.54  & 36.94 & \textbf{54.00} & 63.86 & 59.80\\
QGHC+MUTAN &-&\textbf{82.96}  & 37.16  & 53.88 & \textbf{64.00} & \textbf{60.13} \\
\bottomrule
\end{tabular}
\caption{Comparisons of question answering accuracy of the proposed approach and the state-of-the-art methods on the VQA dataset without using the attention mechanism.}
\label{tab:no_attention}
\end{table}

\subsubsection{Comparison with state-of-the-art methods.}
\label{sssec:vqa_results}

QGHC fuses multi-modal features in an efficient way. The output feature maps of our QGHC module utilize the textual information to guide the learning of visual features and outperform state-of-the-art feature fusion methods. In this section, we compare our proposed approach (without using the attention module) with state-of-the-arts. The results on the VQA dataset are shown in Table \ref{tab:no_attention}. We compare our proposed approach with multi-modal feature concatenation methods including MCB \cite{fukui2016multimodal}, MLB \cite{kim2016hadamard}, and MUTAN \cite{ben2017mutan}. Our feature fusion is performed before the spatial pooling and can better capture the spatial information than previous methods.
Since MUTAN can be combined with MLB (denoted as MUTAN+MLB) to further improve the overall performance.

\begin{table}[tb]
\scriptsize
\centering
\begin{tabular}{l c c c c c}
\toprule
\multirow{2}{*}{Model} &      \multicolumn{4}{c}{test-dev} & \multicolumn{1}{c}{test-std} \\
\cmidrule(lr){2-5} \cmidrule(lr){6-6} 
 & Y/N  &Number & Other & All & All\\
\hline
SMem \cite{xu2016ask} & 80.90 & 37.30  & 43.10  & 58.00 & 58.20    \\
NMN \cite{andreas2016neural}& 81.20 & 38.00 & 44.00 & 58.60 &   58.70    \\
SAN \cite{yang2016stacked}& 79.30 & 36.60 & 46.10 & 58.70 &  58.90 \\     
MRN \cite{kim2016multimodal}& 80.81 & 35.91 & 46.43 & 59.40 & 57.39 \\
DNMN \cite{andreas2016learning}& 81.10 & 38.60 & 45.40 & 59.40 & 59.40 \\
MHieCoAtt \cite{lu2016hierarchical}&79.70 &38.70 & 51.70 & 61.80 & 62.10 \\
MODERN \cite{de2017modulating}& 81.38 & 36.06 & 51.64 & 62.16 & -\\
RAU \cite{noh2016training}& 81.90 & 39.00 & 53.00 & 63.30 &   63.20  \\
MCB+Att \cite{fukui2016multimodal}& 82.20 & 37.70 & 54.80 & 64.20 &  -\\
DAN \cite{nam2016dual}&  83.00 & 39.10 & 53.90 & 64.30 &  64.20 \\ 
MFB+Att \cite{yu2017multi}& 82.50  & 38.30  & 55.20  & 64.60 & - \\
EENMN \cite{hu2017learning}&  - & -   & -   &  64.90 &  -\\
MLB+Att \cite{kim2016hadamard}& \textbf{84.02} & 37.90 & 54.77 & 65.08 & 65.07 \\
MFB+CoAtt \cite{yu2017multi}& 83.20 & \textbf{38.80} & 55.50 & 65.10 &   -     \\
\hline
QGHC+Att+Concat &  83.54  &  38.06   &  \textbf{57.10}     &  \textbf{65.89}   &   \textbf{65.90} \\
\bottomrule
\end{tabular}
\caption{Comparisons of question answering accuracy of the proposed approach and the state-of-the-art methods on the VQA dataset with the attention mechanism.}
\label{tab:vqa}
\end{table}

Attention mechanism is widely utilized in VQA algorithms for associating words with image regions.
Our method can be combined with attention models for predicting more accurate answers. In Section \ref{ssec:attention}, we adopt a simple attention implementation. More complex attention mechanisms, such as hierachical attention \cite{lu2016hierarchical} and stacked attention \cite{yang2016stacked} can also be combined with our approach. 
The results in Table \ref{tab:vqa} list the answering accuracies on the VQA dataset of different state-of-the-art methods with attention mechanism.

We also compare our method with dynamic parameter prediction methods. DPPNet \cite{noh2016image} (Table \ref{tab:no_attention}) and MODERN \cite{de2017modulating} (Table \ref{tab:vqa}) are two state-of-the-art dynamic learning methods.
Compared with DPPNet(VGG) and MODERN(ResNet-152), QGHC improves the performance by 6.78\% and 3.73\% respectively on the test-dev subset, which demonstrates the effectiveness of our QGHC model.

\subsection{CLEVR dataset}

The CLEVR dataset \cite{johnson2016clevr} is proposed to test the reasoning ability of VQA tasks, such as counting, comparing, and logical reasoning. 
Questions and images from CLEVR are generated by a simulation engine that randomly combines 3D objects. This dataset contains 699,989 training questions, 149,991 validation questions, and 149,988 test questions. 

\subsubsection{Experimental setting.}

In our proposed model, the image is resized to $224\times 224$. The question is first embedded to a 300-d vector through a FC layer followed by a ReLU non-linear function, and then input into a 2-layer LSTM with 256 hidden states to generate textual features. Our QGHC network contains three QGHC modules for fusing multi-modal information. All parameters are learned from scratch and trained in an end-to-end manner. The network is trained using the ADAM optimizer with the learning rate $5\times 10^{-4}$ and batch size 64. All the results are reported on the validation subset.

\begin{figure*}[t!]
 \footnotesize
 \begin{tabular}{c c}
  &
  \begin{tabular}{c p{1.15in} p{1.15in} p{1.15in}}
   
   \includegraphics[width=1.15in]{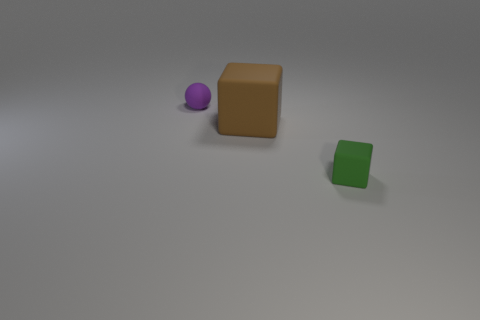}&
   \includegraphics[width=1.15in]{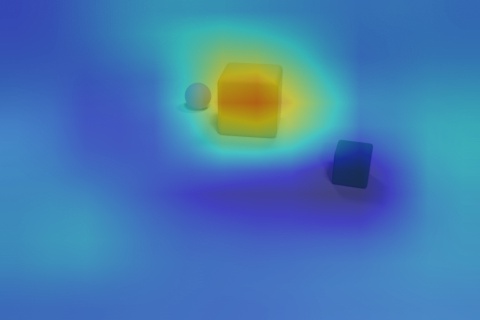}&
   \includegraphics[width=1.15in]{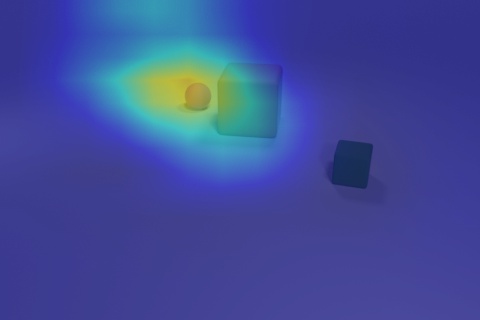}&
   \includegraphics[width=1.15in]{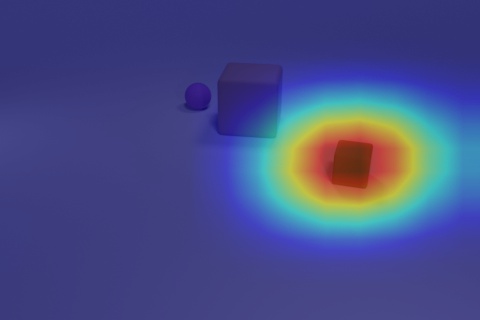}\\
         & 
   \textbf{Q}: {What shape is the \underline{\emph{yellow}} thing?} & 
   \textbf{Q}: {What shape is the \underline{\emph{purple}} thing?} & 
   \textbf{Q}: {What shape is the \underline{\emph{green}} thing?}\\
         & 
   \textbf{A}: {cube} & 
   \textbf{A}: {sphere} & 
   \textbf{A}: {cube}\\ 
   
   \includegraphics[width=1.15in]{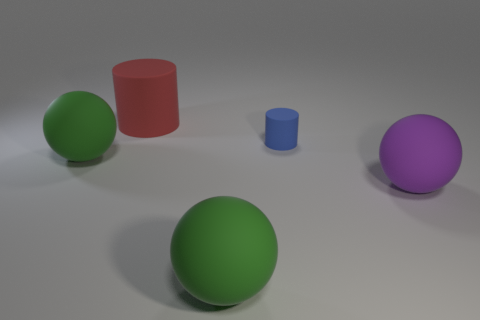}&
   \includegraphics[width=1.15in]{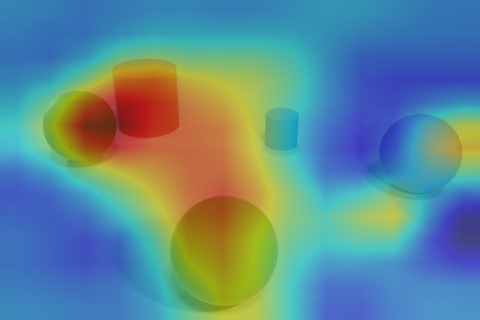}&
   \includegraphics[width=1.15in]{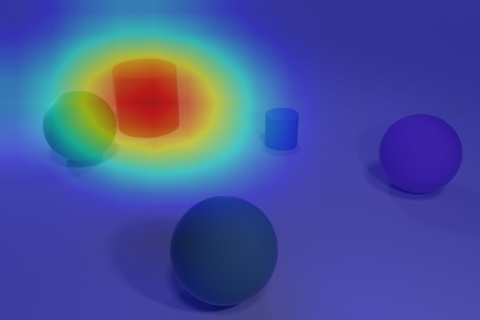}&
   \includegraphics[width=1.15in]{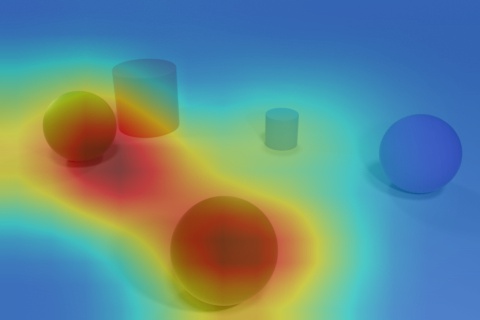}\\ 
         & 
   \textbf{Q}: {What \underline{\emph{number}} of things are rubber in front of the tiny matte cylinder or 
                big purple things} & 
   \textbf{Q}: {The large cylinder that is the same 
                material as the purple is what \underline{\emph{color}}?} & 
   \textbf{Q}: {How many \underline{\emph{green}} things?}\\
       & 
   \textbf{A}: {3} & 
   \textbf{A}: {Red} & 
   \textbf{A}: {2}
  \end{tabular}
 \end{tabular}
 \caption{Visualization of answer activation maps generate by the QGHC.}
 \label{fig:vis}
\end{figure*}

\subsubsection{Comparison with state-of-the-arts.}

We compare our model with the following methods. \emph{CNN-LSTM} \cite{antol2015vqa} encodes images and questions using CNN and LSTM respectively. The encoded image features and question features are concatenated and then passed through a MLP to predict the final answers. \emph{Multimodal Compact Bilinear Pooling (MCB)} \cite{fukui2016multimodal} fuses textual and visual feature by compact bilinear pooling which captures the high level interaction between images and questions.
\emph{Stacked Attention (SA)} \cite{yang2016stacked} adopts multiple attention models to refine the fusion results and utilizes linear transformations to obtain the attention maps. MCB and SA could be combined with the above CNN-LSTM method.
\emph{Neural Module Network (NMN)} \cite{andreas2016neural} propose a sentence parsing method and a dynamic neural network. However, sentence parsing might fail in practice and lead to bad network structure.
\emph{End-to-end Neural Module Network (N2NMN)} \cite{hu2017learning} learns to parse the question and predicts the answer distribution using dynamic network structure.

The results of different methods on the CLEVR dataset are shown in Table \ref{tab:clver}. The multi-modal concatenation (CNN-LSTM) does not perform well, since it cannot model the complex interactions between images and questions. Stacked Attention (+SA) can improve the results since it utilizes the spatial information from input images. Our QGHC model still outperforms +SA by 17.40\%. 
For the N2NMN, it parses the input question to dynamically predict the network structure. Our proposed method outperforms it by 2.20\%.

\begin{table*}[tb]
\centering
\footnotesize
\resizebox{\textwidth}{15mm}{
\begin{tabular}{l  c   c   c     c c c      c c c c       c c c c}
\toprule 
 & & & & \multicolumn{3}{c}{Compare integers} & \multicolumn{4}{c}{Query attribute} & \multicolumn{4}{c}{Compare attribute}\\
\cmidrule(lr){5-7} \cmidrule(lr){8-11}  \cmidrule(lr){12-15}
Model & Overall & Exist & Count & equal & less & more  & size & color & material & shape & size & color & material & shape\\
\hline
Human  \cite{johnson2017inferring} &  92.60 &    96.60 & 86.70 &    79.00 & 87.00& 91.00 & 97.00 & 95.00 &94.00 & 94.00 & 94.00 & 98.00 & 96.00 & 96.00    \\
\hline
CNN-LSTM \cite{antol2015vqa} &  52.30 &  65.20  & 43.70  & 57.00 & 72.00 & 69.00 & 59.00 & 32.00 & 58.00 & 48.00 & 54.00 & 54.00 & 51.00 & 53.00 \\   
\,\,\, +MCB \cite{fukui2016multimodal} &  51.40 & 63.40 & 42.10 & 57.00 & 71.00 & 68.00 & 59.00 & 32.00 &  57.00 & 48.00 & 51.00 & 52.00 & 50.00 & 51.00 \\
\,\,\, +SA \cite{yang2016stacked} &  68.50    &  71.10  &  52.2 & 60.00 & 82.00 & 74.00 & 87.00 & 81.00 & 88.00 & 85.00 & 52.00 & 55.00 & 51.00 & 51.00     \\
NMN \cite{andreas2016neural} & 72.10 & 79.30 & 52.50 & 61.20 & 77.90 & 75.20 & 84.20 & 68.90 & 82.60 & 80.20 & 80.70 & 74.40 & 77.60 & 79.30  \\
N2NMN \cite{hu2017learning} &  83.70 & 85.70 & 68.50 & 73.80 & 89.70 & 87.70 & 93.10 & 84.50 & 91.50 & 90.60 & 92.60 & 82.80 & 89.60 & 90.00 \\
FiLM \cite{perez2017film} & 97.7 & 99.1 & 94.3 & \multicolumn{3}{c}{96.8} & \multicolumn{4}{c}{99.1} & \multicolumn{4}{c}{99.1}\\
\hline
QGHC(ours) &             86.30 &   78.10  &  91.17 &67.30 &    87.14 &      83.28 & 93.65 & 87.86 & 86.75 & 90.70 & 86.24 & 87.24 & 86.75 &  86.93\\
\bottomrule
\end{tabular}}
\caption{Comparisons of question answering accuracy of the proposed approach and the state-of-the-art methods on the CLVER dataset.}
\label{tab:clver}
\end{table*}

\subsection{Visualization of question-guided convolution}
\label{ssec:visualization}

Motivated by the class activation mapping (CAM) \cite{zhou2015simple}, we visualize the activation maps of the output feature maps generated by the QGHC modules. The weighted summation of the topmost feature maps can localize answer regions.

Convolution activation maps for our last QGHC module are shown in Figure \ref{fig:vis}. We can observe that the activation regions relate to the questions and the answers are predicted correctly for different types of questions, including shape, color, and number. In addition, we also visualize the activation maps of different QGHC modules by training an answer prediction FC layer for each of them.
As examples shown in Figure \ref{fig:QGC}, the QGHC gradually focus on the correct regions.

\section{Conclusion}
\label{sec:conclusion}

In this paper, we propose a question-guided hybrid convolution for learning discriminative multi-modal feature representations. Our approach fully utilizes the spatial information and is able to capture complex relations between the image and question. By introducing the question-guided group convolution kernels with both dynamically-predicted and freely-updated kernels, the proposed QGHC network shows strong capability on solving the visual question answering problem. The proposed approach is complementary with existing feature fusion methods and attention mechanisms. Extensive experiments demonstrate the effectiveness of our QGHC network and its individual components.

\section{Acknowledgement}
This work is supported by SenseTime Group Limited, the General Research Fund sponsored by the Research Grants Council of Hong Kong (Nos. CUHK14213616, CUHK14206114, CUHK14205615, CUHK14203015, CUHK14239816, CUHK419412, CUHK14207814, CUHK14208417, CUHK14202217), the Hong Kong Innovation and Technology Support Program (No.ITS/121/15FX), the National Research Foundation, Prime Minister's Office, Singapore under its International Research Centres in Singapore Funding Initiative.

%
%
%
%

\end{document}